\documentclass[11pt]{article}

\usepackage[preprint]{acl}

\usepackage{times}
\usepackage{latexsym}

\usepackage[T1]{fontenc}
\usepackage[utf8]{inputenc}
\usepackage{microtype}
\usepackage{inconsolata}

\usepackage{graphicx}
\usepackage{booktabs}

\usepackage{amsmath}
\usepackage{amssymb}
\usepackage{mathtools}

\usepackage[capitalize,noabbrev]{cleveref}

\title{SONAR-LLM: Autoregressive Transformer that Thinks in Sentence Embeddings and Speaks in Tokens}

\author{
  \textbf{Nikita Dragunov\textsuperscript{1}},
  \textbf{Temurbek Rahmatullaev\textsuperscript{1}},
  \textbf{Elizaveta Goncharova\textsuperscript{1}},
  \textbf{Nikita Kurdiukov\textsuperscript{2}},
  \\
  \textbf{Aysel Mirzoeva\textsuperscript{3}},
  \textbf{Anna Borisiuk\textsuperscript{4}},
  \textbf{Andrey Kuznetsov\textsuperscript{1}},
  \textbf{Anton Razzhigaev\textsuperscript{1}}
  \\[0.3em]
  \textsuperscript{1}FusionBrain Lab, AXXX \quad
  \textsuperscript{2}T-Tech \quad
  \textsuperscript{3}MSU \quad
  \textsuperscript{4}DS-NLP Group, AXXX
  \\
  \small{\textbf{Correspondence:} \href{mailto:nikitadragunovjob@gmail.com}{nikitadragunovjob@gmail.com}}
}

\begin{document}
\maketitle

\begin{abstract}
The recently proposed Large Concept Model (LCM) generates text by predicting a sequence of sentence-level embeddings and training with either mean–squared error or diffusion objectives. We present \textbf{SONAR-LLM}, a decoder-only transformer that thinks in the same continuous SONAR embedding space yet is supervised through token-level cross-entropy propagated via the frozen SONAR decoder. This hybrid objective retains the semantic abstraction of LCM while eliminating its diffusion sampler and restoring a likelihood-based training signal. Across model sizes from 39 M to 1.3 B parameters, SONAR-LLM attains competitive generation quality. We report scaling trends, ablations, benchmark results and a theoretical analysis of inference FLOPs for context sizes up to 1 million tokens. We release the complete training code and pretrained checkpoints to foster reproducibility and future research.
\end{abstract}

\section{Introduction}
\label{sec:introduction}

\begin{figure}[t]
  \centering
  \includegraphics[width=\columnwidth]{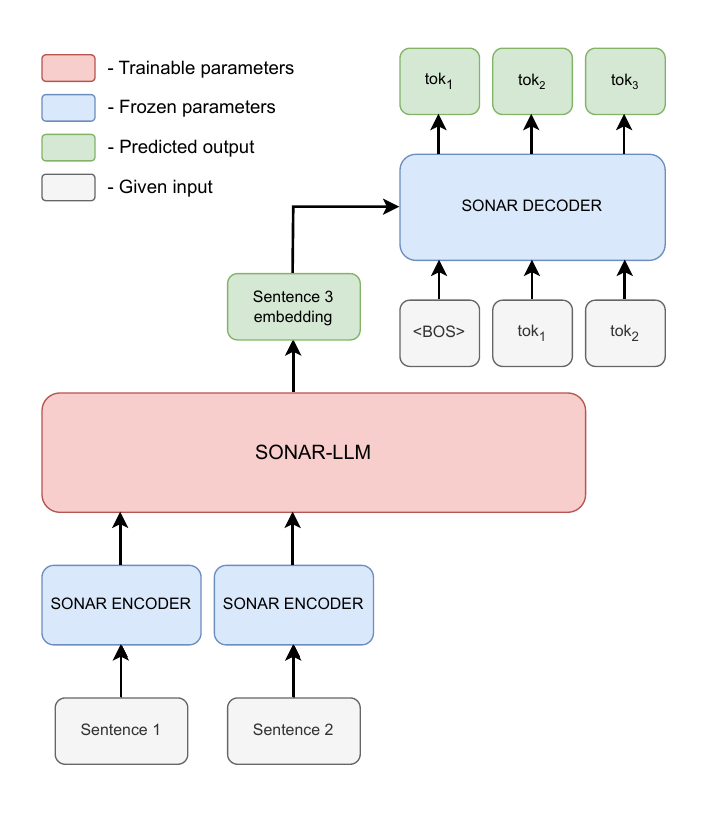}
  \caption{Architecture of SONAR-LLM. The model autoregressively predicts the next sentence embedding given a prefix of embeddings and decodes it via the frozen \textsc{SONAR} decoder.}
  \label{fig:architecture}
\end{figure}

Most autoregressive language models learn token-by-token: they minimize
cross-entropy over a discrete vocabulary and emit one token per forward
step~\cite{Brown2020, t5}.  
This fine-grained decoding is simple to train and
evaluate but becomes a throughput bottleneck for long sequences.  
Meta’s recently introduced Large Concept Model (LCM)~\cite{lcm2024}
addresses the latency issue by predicting a much shorter trajectory of
sentence-level embeddings trained with
diffusion or MSE objective.  
Yet removing token-level likelihoods makes optimization less stable.

We present \textbf{SONAR-LLM}, an autoregressive decoder-only transformer
that keeps LCM’s “think in sentence embeddings” idea while leveraging the
advantages of cross-entropy learning.  
The model predicts SONAR~\cite{sonar2023} sentence embeddings but propagates loss through
the frozen SONAR decoder down to individual tokens, coupling continuous
reasoning with discrete supervision.  
This yields a one-shot sentence generator that is efficient, diffusion-free,
likelihood-consistent, and fast at inference time.

Our contributions are:
\begin{enumerate}
    \item \textbf{Token-Aware Embedding Objective.}  
          We introduce a training objective that back-propagates
          token-level cross-entropy through a frozen SONAR decoder,
          aligning continuous predictions with discrete targets.
    \item \textbf{Scaling Laws Analysis.}
          We provide a detailed scaling law fit for validation losses across model sizes, quantifying the scaling exponents for LLM, LCMs, and SONAR-LLM architectures.
\item \textbf{Text Quality Evaluation.}
      We evaluate the quality of generated texts and primarily compare SONAR-LLM
      against LCM-based sentence-level models using standard NLG metrics
      (BLEU, ROUGE-L, METEOR) and a GPT-4o-based model-as-a-judge evaluation,
      demonstrating consistently higher text quality.
\item \textbf{Summarization Evaluation.}  
      We compare models on summarization tasks using XSum and CNN/DM benchmarks, showing that SONAR-LLM matches or exceeds the performance of other sentence-level approaches.
    \item \textbf{Inference Efficiency Analysis.}
          We provide a theoretical analysis of inference FLOPs for embedding-based sentence-level models, showing that operating on sentence segments yields favorable scaling on long sequences compared to token-level decoding.
    \item \textbf{Reproducible Open-Source Release.}  
          All training code, evaluation scripts, and model checkpoints
          are publicly released to facilitate follow-up research.\footnote{\url{https://github.com/FusionBrainLab/SONAR-LLM}}
\end{enumerate}

\section{Related Work}
\label{sec:related}

\subsection{Token-level autoregressive models.}
Large language models are trained by next-token
prediction with cross-entropy over a discrete vocabulary
\cite{Brown2020}, inheriting the Transformer architecture
\cite{Vaswani2017}.  Recent research has explored alternatives to
self-attention for faster long-sequence processing; for example,
\textsc{Mamba} replaces attention with selective state-space updates
and achieves linear-time generation while matching Transformer quality
\cite{Gu2023Mamba},  with \textsc{Mamba-2} further improving efficiency 
through structured state-space duality \cite{Dao2024Mamba2}. Google's \textsc{Titans} architecture introduces a neural long-term 
memory module that learns to memorize at test time, combining 
attention-based short-term memory with adaptive long-term storage 
to scale beyond 2~M context tokens \cite{Behrouz2025Titans}.

\subsection{Latent-variable text generators.}
Continuous and discrete VAEs generate sentences from latent codes
\cite{Bowman2016}.  Vector-Quantised VAE (VQ-VAE) models compress
sentences into a short sequence of discrete indices and decode them
with an autoregressive prior \cite{VanDenOord2017}.  The
\textbf{SONAR} encoder–decoder extends this idea to a
language-agnostic, multimodal sentence embedding space covering 200
languages \cite{sonar2023}.  Meta’s Large Concept Model
(LCM) builds an autoregressive prior over SONAR embeddings and
investigates MSE, quantization and diffusion losses in that space
\cite{lcm2024}.  Our \textbf{SONAR-LLM} also operates in SONAR
space but reinstates token-level cross-entropy by back-propagating
through the frozen decoder.

\subsection{Diffusion and discrete denoising models for text.}
Diffusion-LM denoises continuous word-embedding sequences to enable
controllable generation without left-to-right constraints
\cite{Li2022DiffusionLM}.  Discrete Denoising Diffusion
Probabilistic Models (D3PMs) corrupt token sequences and learn to
reverse the process in discrete space \cite{Austin2021}.  Recent work
improves training with a score-entropy objective, narrowing the
perplexity gap to autoregressive baselines \cite{Lou2024}. \textsc{LLaDA} demonstrates that masked diffusion models can match 
autoregressive LLMs at scale: an 8B-parameter model trained from scratch 
achieves performance comparable to LLaMA3~8B across diverse benchmarks \cite{Nie2025LLaDA}.

\subsection{Flow and ODE-based generators.}
Flow Matching trains continuous normalizing flows without expensive
simulation and subsumes diffusion as a special case
\cite{Lipman2023FlowMatching}.  Applying flow matching to text,
\textsc{FlowSeq} generates high-quality sentences in a handful of ODE
steps, greatly accelerating sampling \cite{Hu2024Flow}.

In summary, research has progressed from token-wise decoding to latent
concept prediction (LCM), diffusion and flow-based models.  SONAR-LLM
bridges these by learning an autoregressive prior \emph{in} sentence
embedding space while retaining likelihood-based supervision.

\section{SONAR-LLM}
\label{sec:method}

The proposed
\textbf{SONAR-LLM} is an autoregressive decoder–only Transformer that
operates directly in the \textsc{SONAR} sentence-embedding
space while being supervised with token-level cross-entropy. The overall architecture of our approach is illustrated in Fig.~\ref{fig:architecture}.

\subsection{Pre-processing and Sentence Segmentation}
\label{ssec:preproc}

We segment text into small units using the \textit{Punkt} unsupervised sentence tokenizer implemented in \textsc{NLTK}~\cite{Kiss2006}. Each sentence $s_{t}$ is encoded with the frozen multilingual \textsc{SONAR} encoder~\cite{sonar2023}, yielding a fixed-length vector $\mathbf{e}_{t} \in \mathbb{R}^{d}$ ($d{=}1024$ in all experiments). Given a prefix of sentence embeddings $(\mathbf{e}_1, \dots, \mathbf{e}_{t-1})$, the model predicts the embedding $\hat{\mathbf{e}}_{t}$ of the next sentence. This predicted vector is then decoded using the frozen \textsc{SONAR} decoder, and the generated sentence is compared to the true next sentence $s_{t}$, which serves as the training target.

\subsection{Model Architecture}
\label{ssec:arch}

SONAR-LLM is a decoder-only Transformer with the same layer pattern as
Llama 3~\cite{llama3} but an \emph{embedding vocabulary} of size one: the model predicts a
continuous vector rather than a discrete token at each step.  
Formally, given prefix
$\mathbf{e}_{<t}=(\mathbf{e}_{1},\dots,\mathbf{e}_{t-1})$, the network
outputs
$\hat{\mathbf{e}}_{t}=f_{\theta}(\mathbf{e}_{<t})\in\mathbb{R}^{d}$.
We train variants from 39~M to 900~M parameters by scaling width
and depth; all use rotary position encodings and RMS-norm.

\subsection{Cross-Entropy Through the Frozen Decoder}
\label{ssec:loss}

To avoid MSE or diffusion objectives, yet keep likelihood-based training, we
\emph{decode} $\hat{\mathbf{e}}_{t}$ back to token logits with the
frozen SONAR decoder $\mathcal{D}$ using teacher-forcing approach, i.e., generate the next token logits based on predicted sentence embedding $\hat{\mathbf{e}}_{t}$ and previous ground-truth tokens $s_{t, <i}$:
\[
\mathbf{z}_{t, i}=\mathcal{D}(\hat{\mathbf{e}}_{t}, s_{t, <i})\in\mathbb{R}^{|\mathcal{V}|}.
\]
We minimize standard cross-entropy between $\mathbf{z}_{t}$ and the
ground-truth token sequence of sentence $s_{t}$:
\begin{equation}
\label{eq:loss}
\begin{aligned}
\mathcal{L}
  &= -\sum_{t=1}^{T} \log p_{\theta}(s_{t} \mid \mathbf{e}_{<t}) \\
  &= -\sum_{t=1}^{T} \sum_{i=1}^{|s_t|} \log \Bigl( \mathrm{softmax}(\mathbf{z}_{t,i})_{s_{t,i}} \Bigr).
\end{aligned}
\end{equation}
Back-propagation flows through $\mathcal{D}$ keeping SONAR frozen and reducing memory overhead.
Teacher–forcing supplies the ground-truth embedding
$\mathbf{e}_{t}$ at the next time step.

\begin{figure}[t]
  \centering
  \includegraphics[width=\columnwidth]{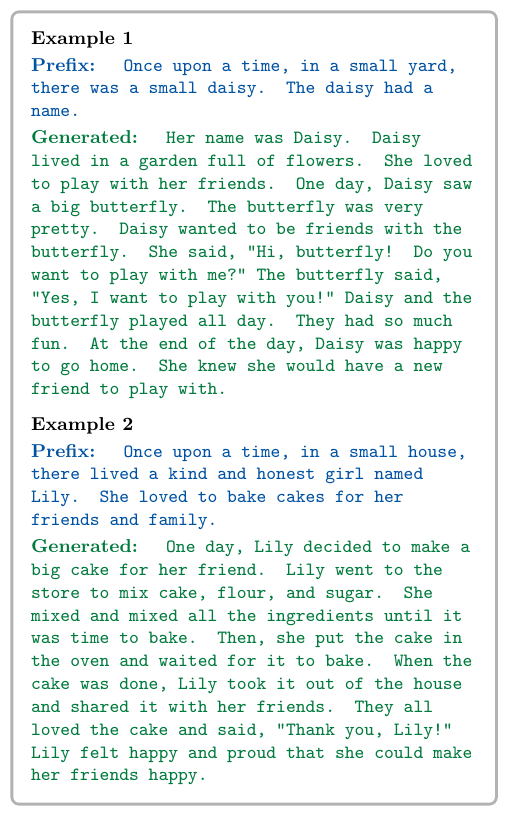}
  \caption{Examples of texts generated by SONAR-LLM 900~M.}
  \label{fig:stories}
\end{figure}

\begin{figure*}[!t]
    \centering
    \includegraphics[width=\linewidth]{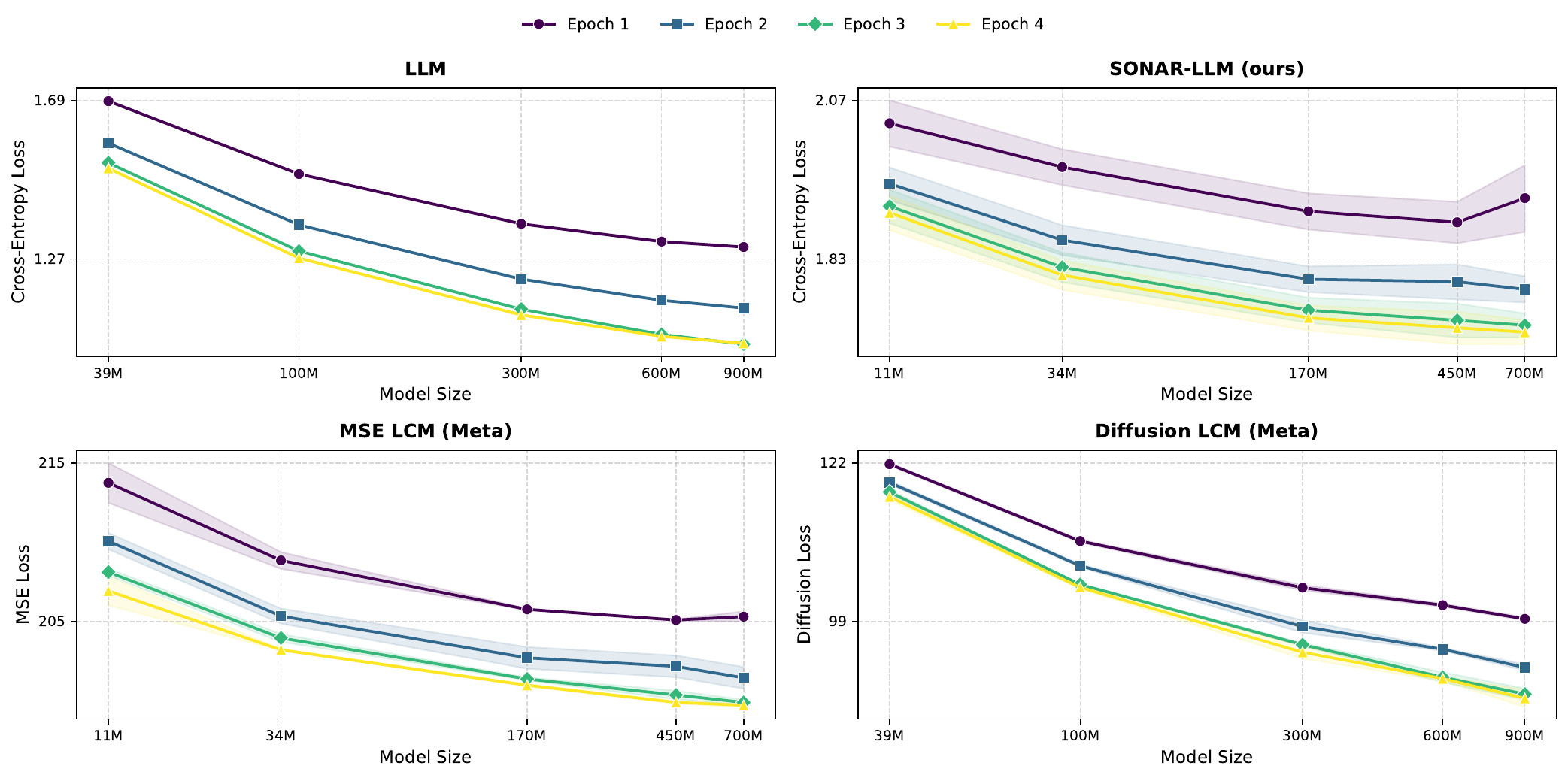}
    \caption{Scaling laws: validation loss dynamics vs. number of trainable parameters.}
    \label{fig:scaling}
\end{figure*}

\begin{figure*}[t]
  \centering
  \includegraphics[width=0.7\linewidth]{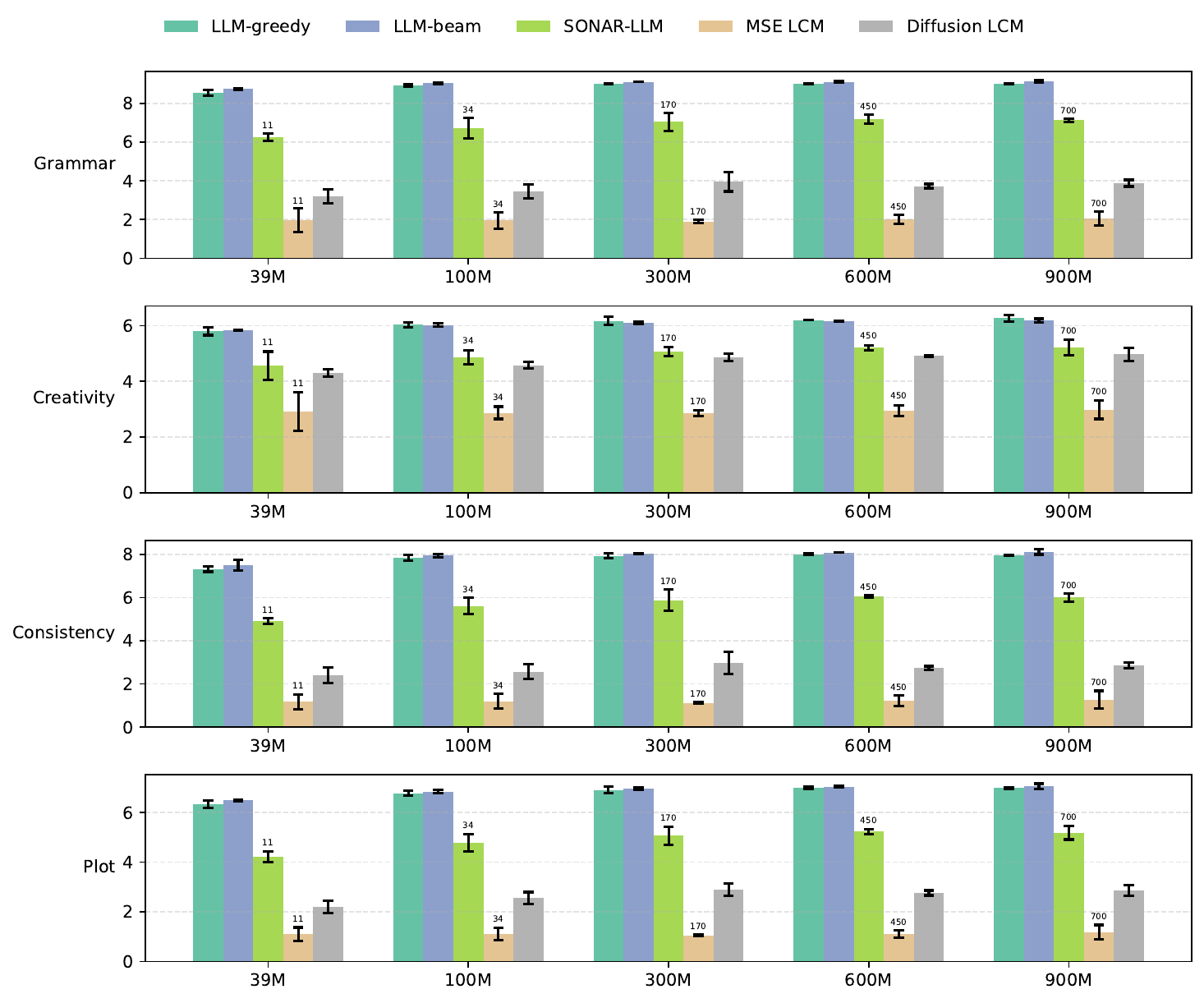}
  \caption{GPT-4o-based evaluation scores (grammar, creativity, consistency, plot) by model and size. Trainable parameter counts are shown above bars for SONAR-LLM and MSE LCM.}
  \label{fig:gpt}
\end{figure*}

\subsection{End of sequence}
\label{ssec:sentinel}

We append a special literal sentence \texttt{"End of sequence."} to
every document and encode it once with the SONAR encoder to obtain
$\mathbf{e}_{\text{eot}}$.  
At inference, generation halts when the cosine similarity between the
latest predicted embedding and $\mathbf{e}_{\text{eot}}$ exceeds
$\tau_{\text{stop}}{=}0.98$, or when $T_{\max}=32$ sentences are
produced.

\section{Results}

We trained large language models (LLMs) of five different scales (39~M, 100~M, 300~M, 600~M, and 900~M parameters) for four epochs each, using the Llama 3 architecture on the \textsc{TinyStories} dataset~\cite{ts2023}. Each run was conducted on a server equipped with up to 8 NVIDIA A100 GPUs (80~GB). When reporting model sizes for LLMs, we included the embedding matrices in the parameter list, as these were fully trained. We also trained SONAR-LLM, MSE-based LCM, and diffusion-based LCM. For SONAR-LLM and MSE-based LCM models, we used the same architecture configurations as their LLM counterparts, but excluded the embedding and decoder parameters from training. As a result, these models contain fewer trainable parameters: 11~M, 34~M, 170~M, 450~M, and 700~M, respectively, having the same depth and width. For consistency, we refer to model sizes (39~M – 900~M) based on the full LLM configurations, even when the number of trainable parameters is smaller. 

For the diffusion-based LCM, we employed the two-tower architecture from the original paper. Both LCM versions were trained using the official implementation provided by the authors~\cite{lcm2024}.

For each model configuration, we performed three independent training runs with three
different random seeds, and report results averaged across these runs.

All models were trained using a cosine learning rate scheduler. We experimented with two learning rates: $5\times10^{-4}$ and $1\times10^{-3}$. Based on validation loss performance, we found $1\times10^{-3}$ to be optimal for SONAR-LLM, while the other models (LLM, MSE-based LCM, and diffusion-based LCM) performed better with a learning rate of $5\times10^{-4}$.

Examples of generated texts can be found in \cref{fig:stories}.

\subsection{Scaling laws}

\begin{figure*}[t]
  \centering
  \includegraphics[width=0.85\linewidth]{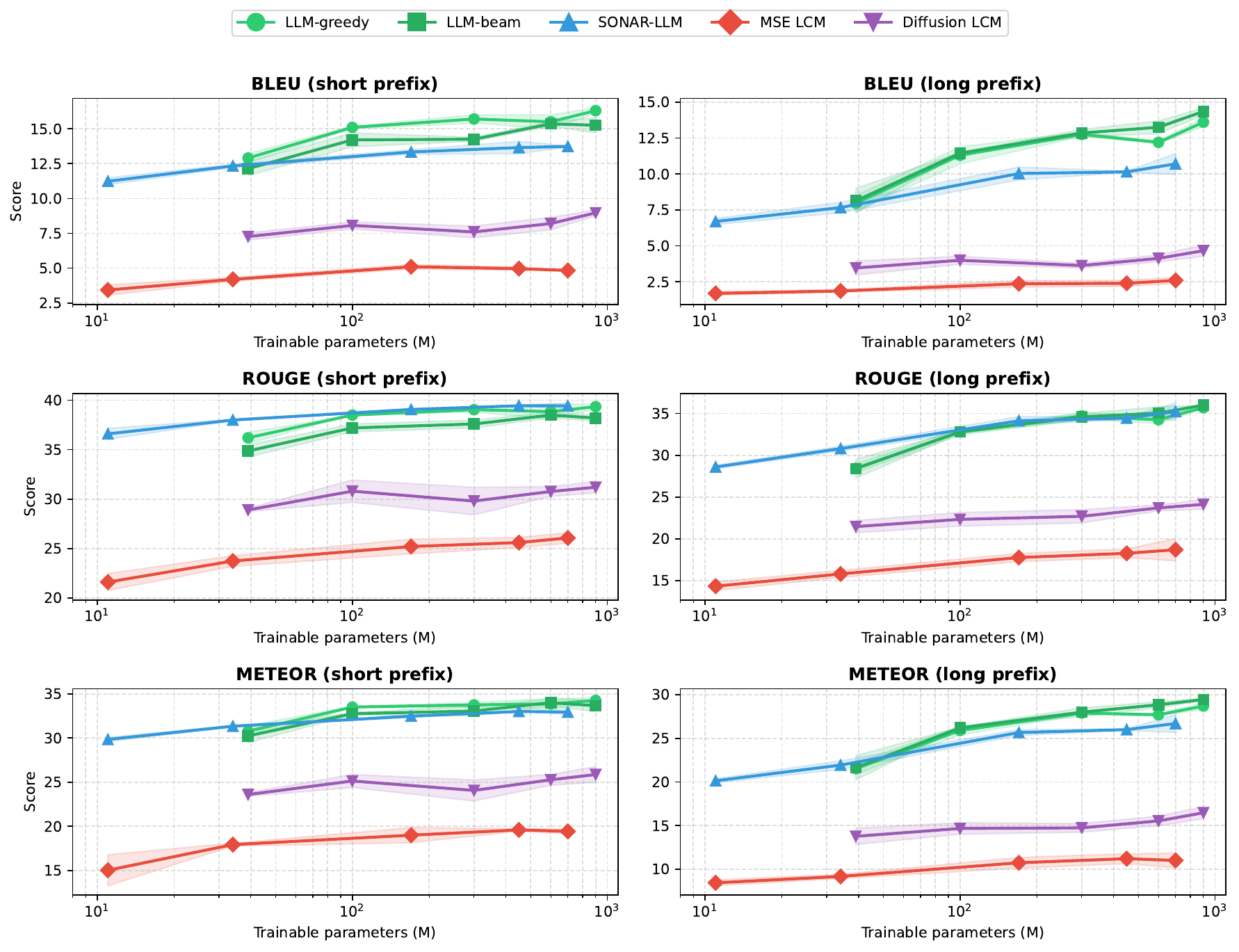}
  \caption{NLG scores by model and size.}
  \label{fig:nlg}
\end{figure*}

The empirical scaling properties of the evaluated architectures, illustrated in Fig.~\ref{fig:scaling}, offer insights into their efficiency in leveraging increased model parameters and training compute. This analysis focuses on the implications of these observed validation loss dynamics for each model type.

We fitted the classical scaling law $$L(N) = a N^{-\alpha} + b$$ to the validation losses of all models at epoch 4. For each architecture, the fit was performed jointly over all model sizes and
three independent random seeds, resulting in 15 data points per model. The results (Table~\ref{tab:scaling_fit}) confirm that SONAR-LLM achieves a strong
scaling exponent ($\alpha \approx 0.596$), comparable to other embedding-based models. For all models, the scaling laws exhibit a good fit to the data, with an $R^2$-score exceeding $0.97$. This demonstrates that SONAR-LLM can efficiently leverage increased model capacity, benefiting from both semantic abstraction and effective scaling behavior.

\begin{table}[t]
\centering
\small
\caption{Fitted scaling law parameters for each model at epoch 4.}
\begin{tabular}{lccc}
\toprule
Model & $a$ & $\alpha$ & $b$ \\
\midrule
LLM & $1.44 \times 10^5$ & 0.724 & 1.04 \\
\textbf{SONAR-LLM (ours)} & $2.86 \times 10^3$ & 0.596 & 1.70 \\
MSE LCM (Meta) & $3.71 \times 10^4$ & 0.525 & 199 \\
Diffusion LCM (Meta) & $2.70 \times 10^5$ & 0.515 & 84.1 \\
\bottomrule
\end{tabular}
\label{tab:scaling_fit}
\end{table}

\subsection{Automatic Evaluation with GPT-4o}

We evaluated the performance of all four model types on a dataset consisting of 512 generated stories, assessing grammatical correctness, creativity, coherence, and plot consistency, following the methodology proposed by~\cite{ts2023}. To initiate story generation, we used the first two sentences from validation set stories as prompts. During evaluation, GPT-4o was shown the full story--including the prompt and the generated continuation--but was explicitly instructed to assess only the continuation starting from the third sentence. All models were evaluated after four epochs of training. For the LLM, we experimented with both greedy decoding and beam sampling with four beams.

As illustrated in Fig.~\ref{fig:gpt}, error bars correspond to three standard
deviations ($3\sigma$), providing $99.7\%$ confidence intervals for the mean scores
across random seeds.
For SONAR-LLM and MSE-based LCM models, the numbers shown above the bars indicate
the corresponding counts of trainable parameters.

The classic token-level LLM demonstrates the best performance across all metrics. With beam sampling, the largest 900 M model achieves strong grammatical correctness (9.1), coherence (8.1), and plot consistency (7.0), with creativity scores around 6.2. Greedy decoding shows comparable results with slightly lower coherence.

Among the concept-based models, our proposed SONAR-LLM achieves the highest story generation quality, significantly outperforming both the diffusion-based and MSE-based LCM variants across all four metrics. The 900 M SONAR-LLM reaches grammatical correctness of 7.1, creativity of 5.2, coherence of 6.0, and plot consistency of 5.2. In contrast, Diffusion LCM shows moderate performance (grammar 3.9, creativity 5.0, coherence 2.9, plot 2.8), while MSE LCM exhibits substantially lower scores across all metrics (grammar 2.1, creativity 3.0, coherence 1.3, plot 1.2), suggesting that direct MSE regression in embedding space struggles to maintain narrative structure.
\subsection{NLG Metrics}

To assess how effectively models capture the distribution of the original data, we evaluated standard NLG metrics, including BLEU, ROUGE-L, and METEOR. Specifically, we selected 512 stories from the validation set and used the first two sentences from each story as a context (short prefix) to generate the third sentence. We then measured similarity between the generated sentence and the corresponding reference sentence from the validation set using the aforementioned metrics. Additionally, we performed the same evaluation using half of each story in terms of sentence count as a context (long prefix), to investigate model performance under varying context lengths. Results are provided in Fig.~\ref{fig:nlg}, where shaded regions represent one standard deviation across random seeds.

The NLG evaluation demonstrates that SONAR-LLM approaches the performance of standard
autoregressive LLMs across most metrics at the largest evaluated scale.
For the 900~M-parameter models (with approximately 700~M trainable parameters for
SONAR-LLM), SONAR-LLM matches LLM performance on ROUGE-L with short prefixes
(39.4 vs.\ 39.4) and achieves competitive results on METEOR (32.9 vs.\ 34.2).
With longer context, similar patterns hold: SONAR-LLM attains comparable ROUGE-L
(35.2 vs.\ 35.7).
In contrast, both MSE-based and diffusion-based LCMs exhibit substantially lower
performance across all metrics at this scale, with MSE LCM achieving 26.1 ROUGE-L
and Diffusion LCM reaching 31.2 ROUGE-L on short prefixes.

\subsection{Summarization Evaluation}
\label{ssec:summarization}

Summarization is a vital benchmark for sentence-level language models, as it directly assesses their capability to capture semantic content and produce coherent, structured text. Prior works on sentence-level LLMs, including the original LCM paper, emphasized summarization as a crucial test of their abstraction and compression abilities. Motivated by this, we evaluated SONAR-LLM and relevant baselines on standard abstractive summarization benchmarks.

We pretrained 1.3~B-parameter models (1.1~B trainable parameters for SONAR-LLM and MSE LCM, excluding embedding matrix) on a diverse mixture of datasets, including \textsc{TinyTextbooks}, \textsc{TinyOrcaTextbooks}, \textsc{TinyStrangeTextbooks}, \textsc{TextbooksAreAllYouNeed}~\cite{textbooks}, 
\textsc{Wikitext-103-detokenized}~\cite{wikitext}, 
\textsc{XSum}~\cite{xsum}, 
\textsc{CNNDailyMail}~\cite{cnn}. We then evaluated summarization performance on test examples from the \textsc{XSum} and \textsc{CNNDailyMail} datasets, generating the same number of sentences as in the reference summaries (typically one sentence for XSum and three sentences for CNN/DM). Results were measured using ROUGE-L and METEOR metrics.

The results in Table~\ref{tab:summarization} indicate that SONAR-LLM substantially outperforms existing sentence-level baselines (MSE LCM and Diffusion LCM) on both datasets, confirming its effectiveness for summarization tasks. Compared to token-level LLMs, SONAR-LLM achieves comparable performance on the more abstractive XSum dataset but remains behind on CNN/DM, which tends to favor more extractive approaches. These observations indicate that SONAR-LLM can be a promising approach for sentence-level tasks involving abstraction and semantic compression.

\begin{table}[t]
\centering
\small
\setlength{\tabcolsep}{2pt}
\caption{Summarization results on XSum and CNN/DM. Mean (std) across seeds.}
\begin{tabular}{lcc|cc}
\toprule
\textbf{Model} & \multicolumn{2}{c}{\textbf{XSum}} & \multicolumn{2}{c}{\textbf{CNN/DM}} \\
               & R-L & MET & R-L & MET \\
\midrule
LLM-greedy &
\textbf{20.4}{\scriptsize (1.1)} &
16.5{\scriptsize (1.1)} &
\textbf{19.3}{\scriptsize (0.6)} &
15.2{\scriptsize (1.1)} \\
LLM-beam &
20.2{\scriptsize (1.1)} &
\textbf{16.9}{\scriptsize (1.1)} &
18.4{\scriptsize (0.1)} &
\textbf{16.9}{\scriptsize (0.4)} \\
SONAR-LLM (ours) &
18.5{\scriptsize (0.7)} &
14.5{\scriptsize (0.5)} &
15.5{\scriptsize (0.7)} &
10.0{\scriptsize (0.4)} \\
Diffusion LCM (Meta) &
12.0{\scriptsize (0.2)} &
8.8{\scriptsize (0.4)} &
10.1{\scriptsize (0.1)} &
4.8{\scriptsize (0.2)} \\
MSE LCM (Meta) &
11.6{\scriptsize (0.5)} &
8.5{\scriptsize (0.2)} &
7.8{\scriptsize (0.2)} &
3.8{\scriptsize (0.1)} \\
\bottomrule
\end{tabular}
\label{tab:summarization}
\end{table}

\subsection{Unfreezing the SONAR Decoder}
\label{ssec:unfreeze_decoder}

In all main experiments, we keep the \textsc{SONAR} encoder and decoder frozen and
train only the autoregressive prior in the sentence-embedding space, primarily for
computational efficiency.

We additionally explored unfreezing the \textsc{SONAR} decoder when training
1.3~B-parameter models on the mixed pretraining corpus described in
\cref{ssec:summarization}.
Unfreezing the decoder consistently reduced validation cross-entropy by more than
10\% across runs.
However, this improvement did not translate into statistically significant gains in
downstream text quality: summarization performance on XSum and CNN/DailyMail remained comparable to those
obtained with a frozen decoder.

We further evaluated this setting on the \textsc{RULER} benchmark~\cite{ruler2024}, focusing on the
NIAH (needle-in-a-haystack) task that requires exact retrieval of numerical values
from long contexts.
For this experiment, we fine-tuned the pretrained 1.3~B-parameter models from
\cref{ssec:summarization} on the RULER dataset.
In this regime, unfreezing the decoder substantially improved exact-match accuracy
(from $21.6\%$ to $41\%$), indicating a clear benefit for symbol-heavy tasks.

Overall, these results suggest that decoder unfreezing is unnecessary for standard
natural-language generation, where the \textsc{SONAR} embedding space is already
well aligned with textual data.
In contrast, for tasks that require precise manipulation or reproduction of
numerical or symbolic content--such as long-context retrieval, mathematics, or
code--adapting the embedding-to-token interface by unfreezing the decoder can be
beneficial.

\subsection{Inference Efficiency}
\label{ssec:inference}

We compared the theoretical inference complexity in FLOPs of SONAR-LLM and a standard LLM depending on the input sequence length. The comparison was performed for models with identical architectures configured at 600~M parameters. In the case of SONAR-LLM, we assumed an average sentence length of 60 tokens and, in addition to the complexity of the main SONAR-LLM model, we also included the FLOPs of the SONAR encoder and decoder. The inference setup of SONAR-LLM follows the same structural principles as the MSE-based LCM proposed by~\cite{lcm2024}, suggesting that both models exhibit similar inference efficiency due to similar design.

\begin{figure}[!t]
    \centering
    \includegraphics[width=\linewidth]{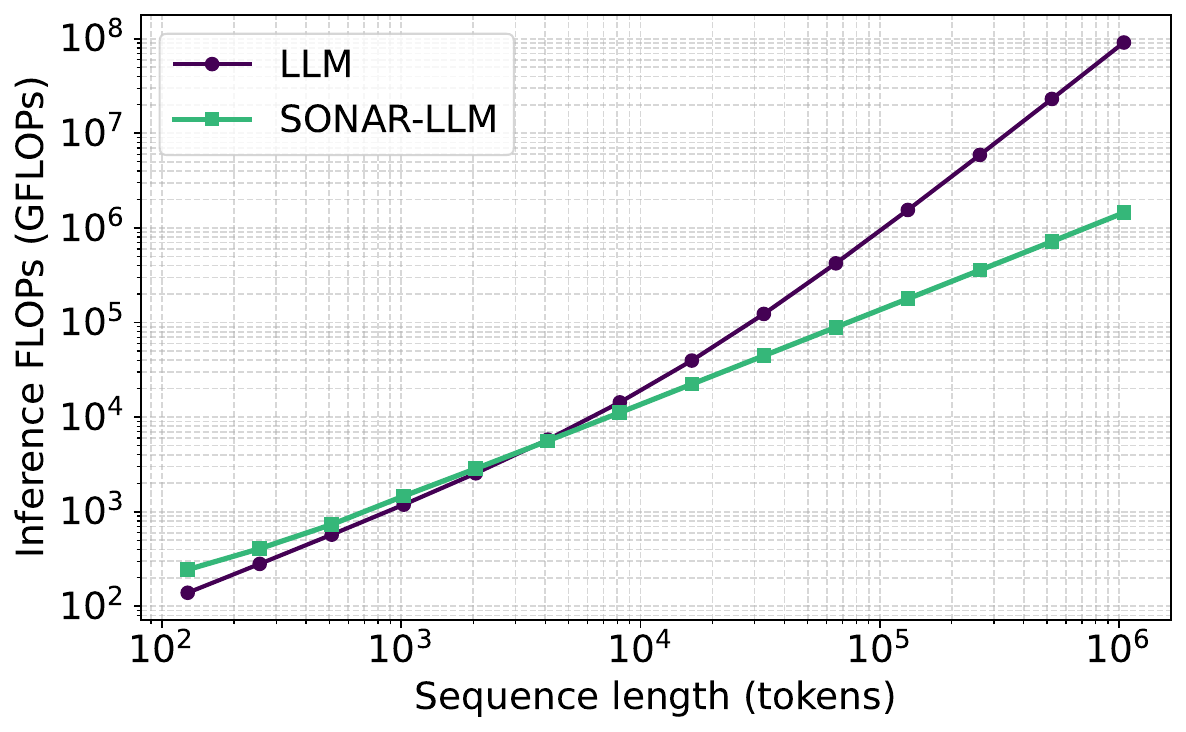}
    \caption{Theoretical inference FLOPs for autoregressive LLM and SONAR-LLM as a function of sequence length (log--log scale).}
    \label{fig:flops}
\end{figure}

The results presented in Fig.~\ref{fig:flops} indicate that, for shorter sequences, standard token-level LLMs maintain a computational advantage due to their optimized token-wise autoregressive decoding. However, as the input length increases, this advantage diminishes: starting from approximately 4096 tokens, SONAR-LLM surpasses the standard LLM in inference efficiency. This is attributable to SONAR-LLM’s design, which processes entire sentences as atomic units, thereby reducing the number of required decoding steps relative to token-based models. While the theoretical computational complexity remains quadratic for both approaches, the effective cost for SONAR-LLM grows much more slowly with sequence length because it operates on a compressed sequence of sentence embeddings. In practice, this yields an almost linear growth in FLOPs up to 1 million tokens, as the quadratic term is scaled by the inverse square of the average sentence length.

\section{Conclusion}
\label{sec:conclusion}
We presented \textbf{SONAR-LLM}, a decoder-only Transformer that predicts sentence embeddings and is supervised via token-level cross-entropy propagated through a frozen SONAR decoder. This approach retains the semantic abstraction of concept-based models like LCM while restoring a likelihood-based training signal.

As a proof of concept, we trained SONAR-LLM on the \textsc{TinyStories} dataset. Across model sizes, SONAR-LLM exhibits stable optimization behavior and competitive scaling trends, matching or improving upon those observed for both MSE-based and diffusion-based LCMs. In GPT-4o evaluations, SONAR-LLM outperformed both LCM variants in grammar, coherence, creativity, and plot consistency. On standard NLG metrics, \textsc{SONAR-LLM} demonstrated strong performance, consistently matching the standard token-level LLM. It also outperformed both the MSE-based and diffusion-based LCMs across all prefix lengths, establishing it as a promising alternative for sentence-level generation tasks.

To broaden the evaluation scope, we pretrained all models on a diverse mixture of instructional and open-domain corpora. This enabled us to assess summarization capabilities on standard datasets such as XSum and CNN/DM. SONAR-LLM achieved consistently stronger performance than prior sentence-level baselines and demonstrated its ability to handle summarization tasks with competitive quality, further validating the effectiveness of our proposed objective in more realistic settings.

Finally, we examined the effect of unfreezing the frozen SONAR decoder.
While decoder unfreezing consistently improves cross-entropy during pretraining,
it does not lead to measurable gains on standard natural-language benchmarks such as summarization.
However, on symbol-heavy long-context retrieval tasks (RULER/NIAH), unfreezing the decoder yields a substantial improvement in exact-match accuracy, nearly doubling performance.
These results indicate that decoder adaptation is unnecessary for semantically grounded text,
but becomes important when precise numeric or alphanumeric reproduction is required.

Our theoretical FLOPs analysis further demonstrates that SONAR-LLM achieves superior inference efficiency for long contexts compared to token-level LLMs: beyond 4096 tokens, its total computational cost grows almost linearly with sequence length up to 1 million tokens. Importantly, this effect results from operating on sentence-level segments, but the underlying complexity is still quadratic. This property enables such embedding-based architectures to serve as a practical and scalable option for long-context generation.

\section*{Limitations}

While our study reveals clear trends among the evaluated model architectures, several limitations remain.

First, our evaluation of generation quality combines standard automatic metrics (BLEU, ROUGE-L, METEOR) with GPT-4o-based assessments of grammar, coherence, creativity, and plot consistency. While the latter offers a stronger proxy for human judgment, it is still limited by the behavior and biases of the underlying model. A more complete evaluation would benefit from direct human annotation or broader qualitative analysis.

Second, our experiments are limited to English-language data, and the reported
results may not directly generalize to other languages.
At the same time, the underlying \textsc{SONAR} encoder--decoder is trained as a
multilingual model and produces language-agnostic sentence embeddings, which
suggests that the observed trends may extend beyond English.
A thorough multilingual evaluation is left for future work.

\section*{Impact Statement}

This paper introduces a sentence-level approach to language generation that
operates in a continuous embedding space while retaining token-level likelihood
training.
By predicting and processing text at the granularity of sentences rather than
individual tokens, the proposed method enables more efficient handling of long
contexts and offers an alternative abstraction for language modeling.
The primary goal of this work is to advance research in efficient, scalable
language generation and representation learning.

While the model reasons in sentence embeddings instead of tokens, it does not
introduce fundamentally new generative capabilities compared to existing large
language models.
The types of outputs it can produce, and the potential misuse scenarios (e.g.,
generation of misleading or low-quality content), remain largely the same as
those associated with standard LLMs.
At the same time, operating at a higher level of abstraction may amplify or
suppress certain properties of text (such as verbosity or structural coherence),
which could affect downstream applications in subtle ways.

Overall, we do not identify novel ethical risks specific to this approach beyond
those already present in current language models.
We expect that the primary impact of this work will be to support research on
efficient long-context modeling, abstraction in language generation, and
architectures that reason over larger semantic units than tokens.

\bibliography{custom}

\end{document}